\documentclass{article}

\usepackage{microtype}
\usepackage{graphicx}
\usepackage{subfigure}
\usepackage{booktabs} % for professional tables

\usepackage{amssymb}
\usepackage{amsmath}
\DeclareMathOperator*{\argmax}{argmax}
\usepackage{multirow}

\usepackage{hyperref}

\usepackage[accepted]{icml2021}

\icmltitlerunning{Mixed Cross Entropy Loss for Neural Machine Translation}

\begin{document}

\twocolumn[
\icmltitle{Mixed Cross Entropy Loss for Neural Machine Translation}



\icmlsetsymbol{equal}{*}

\begin{icmlauthorlist}
\icmlauthor{Haoran Li}{equal,sutd}
\icmlauthor{Wei Lu}{equal,sutd}
\end{icmlauthorlist}

\icmlaffiliation{sutd}{StatNLP Research Group, Singapore University of Technology and Design, Singapore}
% \icmlaffiliation{to}{Department of Computation, University of Torontoland, Torontoland, Canada}
% \icmlaffiliation{goo}{Googol ShallowMind, New London, Michigan, USA}
% \icmlaffiliation{ed}{School of Computation, University of Edenborrow, Edenborrow, United Kingdom}

\icmlcorrespondingauthor{Wei Lu}{luwei@sutd.edu.sg}
% \icmlcorrespondingauthor{Cieua Vvvvv}{c.vvvvv@googol.com}
% \icmlcorrespondingauthor{Eee Pppp}{ep@eden.co.uk}

\icmlkeywords{Neural Machine Translation, Conditional Language Generation, Exposure Bias}

\vskip 0.3in
]



\printAffiliationsAndNotice{\icmlEqualContribution} % otherwise use the standard text.

\begin{abstract}
In neural machine translation, cross entropy  (CE) is the standard loss function in two training methods of auto-regressive models, i.e., \emph{teacher forcing} and \emph{scheduled sampling}.
In this paper, we propose \emph{mixed cross entropy loss} (mixed CE) as a substitute for CE in both training approaches.
In teacher forcing, the model trained with CE regards the translation problem as a one-to-one mapping process, while in mixed CE this process can be relaxed to one-to-many.
In scheduled sampling, we show that mixed CE has the potential to encourage the training and testing behaviours to be similar to each other, more effectively mitigating the \emph{exposure bias} problem.
We demonstrate the superiority of mixed CE over CE on several machine translation datasets, WMT'16 Ro-En, WMT'16 Ru-En, and WMT'14 En-De in both teacher forcing and scheduled sampling setups.
Furthermore, in WMT'14 En-De, we also find mixed CE consistently outperforms CE on a multi-reference set as well as a challenging paraphrased reference set.
We also found the model trained with mixed CE is able to provide a better probability distribution defined over the translation output space. {Our code is available at \url{https://github.com/haorannlp/mix}}.
\end{abstract}

\section{Introduction}
\label{motiv}
Conditional language generation tasks, e.g., machine translation, text summarization, are all text-to-text problems. 
The most popular models to solve these tasks include RNNs \cite{Elman1990FindingSI,Hochreiter1997LongSM,cho-etal-2014-learning}, Transformer \cite{NIPS2017_7181}, etc., which are usually arranged in an encoder-decoder architecture \cite{NIPS2014_5346, cho-etal-2014-learning}. 
Given a training example $(x, y)\in \mathcal{D}$, the encoder first compresses the source sequence $x=(x_1,...,x_m)$ into a vector $h$ and then the decoder will produce a target sequence $\hat{y}=(\hat{y}_1,...,\hat{y}_n)$ from $h$\footnote{With potential reference to $x$, e.g., through the attention mechanism. We ignore this for brevity in our discussion.}. 
However, due to information loss in the compression process and the limited expressiveness power of the model, it can be hard for the decoder to recover a good target sequence $\hat{y}$ from $h$ alone.
To solve this problem, people often use \emph{teacher forcing} \cite{6795228} where both $h$ and the gold target $y=(y_1,...,y_n)$ are fed into the decoder during training.
{Next, the aim is typically to minimize the {\em cross entropy} (CE) loss, which can be written as $-\sum_{t=1}^{n}\log p_{\theta}(y_t|y_{<t},x)$.\footnote{Here $y_{<t}=(y_0,y_1,...,y_{t-1})$ is the partial target sequence, where $y_0$ is the special start token, and $\theta$ denotes model parameters.}}
The general idea is that, by optimizing the CE, we hope the output probability distribution {$p_{\theta}(\cdot|y_{<t},x)\in \mathbb{R}^{|V|}$} ($|V|$ is the vocabulary size) can approximate the target one-hot encoding of $y_t$ {(a vector with only one position being 1, others being 0)}.
In practice, empirical success of various {\color{black}neural machine translation (NMT)} models trained with CE \citep{DBLP:journals/corr/BahdanauCB14, NIPS2017_7181} has demonstrated CE's effectiveness.

\begin{figure}[t!]
%\vskip 0.2in
\begin{center}
\centerline{\includegraphics[width=\columnwidth,]{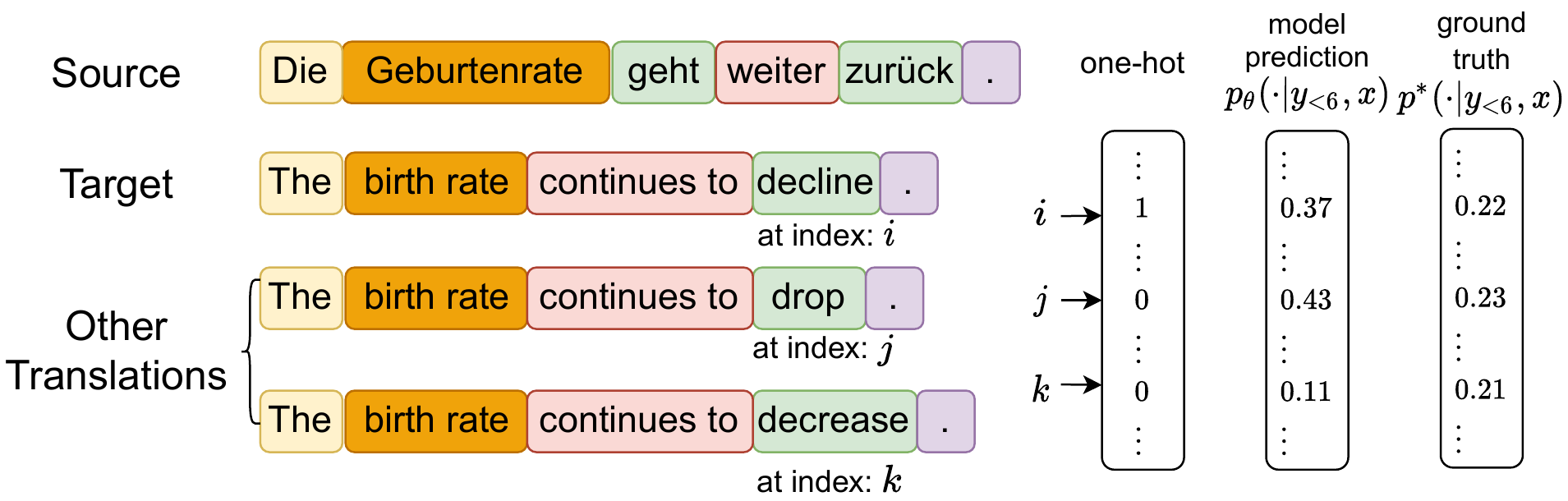}}
\caption{Example of machine translation's one-to-many property. ``Source'' and ``Target'' belong to the training set while ``Other Translations'' denotes other plausible translations which are not present in the training data.}
\label{motiv_fig}
\end{center}
%\vskip -0.2in
\end{figure}

However, it is worth noting that NMT is inherently a {\em one-to-many} mapping problem where a source sentence has multiple plausible translations.
In Fig. \ref{motiv_fig}, suppose we want to predict the last word $y_6$ ``\emph{decline}'' conditioned on the source sentence and the prefix $y_{<6}$.
Although ``\emph{decline}'' is the gold target that our model should assign the most probability mass to, other synonyms of ``\emph{decline}'' such as ``\emph{drop}'', ``\emph{decrease}'' are also plausible translations in this context, namely the corresponding values of these synonyms in the vector representation of $y_6$ should be non-trivial.
Ignoring these synonyms and simply fitting the one-hot encoding may limit the model's generalization ability because $p_{\theta}(\cdot|y_{<t},x)$ deviates from the ground truth $p^*(\cdot|y_{<t}, x)$ where the test data {\color{black}is} drawn \citep{RAML, 44903, xiao2019dual}.
Nevertheless, there is no doubt that when we train our model with CE, the predictions $p_{\theta}(\cdot|y_{<t},x)$ still contain useful information about the ground truth $p^*(\cdot|y_{<t}, x)$.
In this paper, we make a simple assumption: given a well-trained model, if the predicted token (the one with the largest probability given by the model) does not match the ground truth, this token is very likely to be a synonym or part of a synonym of the ground truth.
{This assumption is supported by the fact that in a typical parallel corpus, the same source word can have multiple translations in different training instances.
With a comparable number of occurrences of different translations given the same source word, the model tends to evenly allocate the probability mass to them after learning, thereby having a chance to predict a synonym of the gold token as the most probable one during decoding.}
{Based on this assumption, in teacher forcing, we use mixed CE to incrementally exploit the information regarding synonyms so as to boost model's generalization capability, which can be better demonstrated in a multi-reference test set (see Section \ref{mix_ce_in_tf}).}

Despite its simplicity, teacher forcing suffers from \emph{exposure bias} \citep{DBLP:journals/corr/RanzatoCAZ15}, which refers to the discrepancy that during training time, gold target $y$ is observable by the decoder whereas at test time $y$ is unknown.
Thus, at test time, the model has to sample from its own predictions to decode auto-regressively.
The major solution to exposure bias is to train the model on its own predictions at training stage such that the mismatch between training/testing input distributions can be reduced \citep{Daum2009SearchbasedSP, pmlr-v15-ross11a, Venkatraman2015ImprovingMP, NIPS2015_5956, DBLP:journals/corr/RanzatoCAZ15, DBLP:conf/iclr/BahdanauBXGLPCB17, leblond2018searnn, zhang-etal-2019-bridging}.
One of the those methods is \emph{scheduled sampling} \citep{NIPS2015_5956} (see Section \ref{intro}), which mixes gold input\footnote{The input here refers to {\em decoder input}. Unless otherwise specified, the {\em encoder input} remains the same.} with model-generated input (model predictions) and then maps this mixed input to the gold target.

{We argue that besides making changes to the input distribution during the training time, so that it simulates the {\em test-time inputs}, we can also mitigate exposure bias by simulating the {\em test-time behaviour}, making the model more robust to the difference between training/testing inputs.
That is, we should design our model in such a way that it always produces similar results no matter whether the input is gold or model-generated.
In scheduled sampling, this can be done with mixed CE, which maps the mixed input not only to the gold target, but also to the output which is obtained from the gold input.}

In this paper, we propose mixed CE to substitute CE in both teacher forcing and scheduled sampling training of NMT models.
In teacher forcing, mixed CE guides the model to learn a one-to-many mapping by incrementally exploiting useful synonym information in the model predictions.
In scheduled sampling, mixed CE encourages the model to output similar results regardless of if the model is fed with the gold input or the mixed input which consists of gold target tokens and model predictions.
We demonstrate the effectiveness of mixed CE not only on the standard test set of WMT'16 Ro-En, WMT'16 Ru-En, WMT'14 En-De but also on a multi-reference set \citep{ott2018analyzing}  as well as a challenging paraphrased reference set \citep{freitag-bleu-paraphrase-references-2020}.

\section{Background}
\label{intro}
In this section, we review some background knowledge about \emph{exposure bias} and \emph{scheduled sampling}.
In teacher forcing, the model uses the gold target $y$ as input during training but $y$ is not available at test time.
Hence, at test time, the decoder has to sample from its own predictions $\hat{y}_{t}$ in the last time step $t-1$ and feed the sampled token as input to the current step $t$.
Note that the model has only been trained on the empirical data distribution instead of its own predictions and this may lead to arbitrary decoding decisions at test time.
To alleviate exposure bias, \citet{NIPS2015_5956} proposed scheduled sampling for RNN training.
In scheduled sampling, the model randomly decides to use gold data $y_{t}$ with probability $\epsilon \in [0,1]$ or the model's own prediction $\hat{y}_{t}$ (the one with the largest probability) with probability $1-\epsilon$ as input at each time step $t$.
$\epsilon$ is initialized to a large value close to 1 and then decays as training proceeds (see Section \ref{exp_setup}), which assures the model can converge better and faster.
Consequently, the model is exposed to part of its predictions during training and this reduces the risk of having a bad generation at test time. 

\begin{figure}[t!]
\vskip 0.2in
\begin{center}
\centerline{\includegraphics[width=\columnwidth]{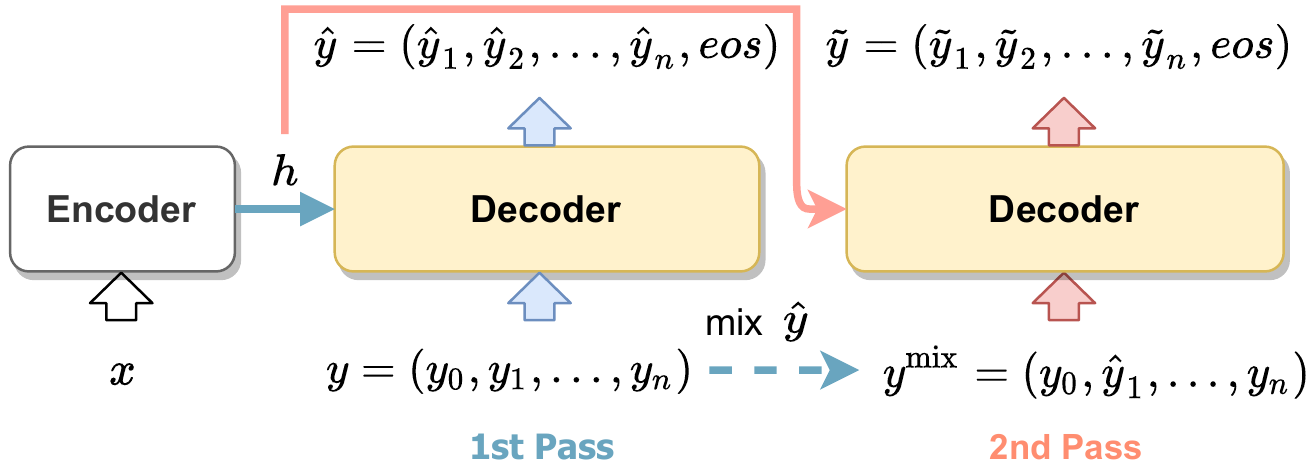}}
\caption{Scheduled sampling for Transformer. We apply $\argmax$ to the output logits, \textcolor{black}{or equivalently the log likelihood $\log p_{\theta}(\cdot|y_{<t},x)$}, at each step {\color{black}$t$} to obtain $\hat{y}$ in the first pass. {{\em eos}} is the end-of-sequence symbol.} %$y^\text{mix}$ is a mix of $(y_1,...,y_n)$ and $(\hat{y}_1,...,\hat{y}_n)$.}
\label{sst}
\end{center}
\vspace{-15pt}
\end{figure}

For Transformer \citep{NIPS2017_7181}, {\color{black}scheduled sampling is typically done in a bit different way} because we want to avoid sequential decoding and make full use of Transformer decoder's parallel computation mechanism ({masked self-attention}) during training.
It has been proposed and widely adopted by the community \cite{mihaylova-martins-2019-scheduled, zhang-etal-2019-bridging, duckworth2020parallel} that we could run the Transformer once without accumulating the gradients and store the greedily predicted sequence $\hat{y}$ for the second pass (see Fig. \ref{sst}).
{Then we randomly replace each token in the gold target sequence $y$ with the corresponding token in the predicted sequence $\hat{y}$ with probability $1-\epsilon$, obtaining a mixed sequence $y^{\text{mix}}$. }
Next we feed $y^{\text{mix}}$ to the Transformer decoder again and compute the loss:
\begin{equation}
\mathcal{L}_{CE}=-\sum_{t=1}^n\log p_{\theta}(y_t|y^{\text{mix}}_{<t},x).
\label{eq0}
\end{equation}
%{\color{red}and be more robust to input variations.}.
Even though the model {involves two forward passes (i.e., perform decoding twice)}, it is still faster than sequential decoding.
{\citet{zhang-etal-2019-bridging} proposed a variant of scheduled sampling called \emph{word oracle}, which adds some Gumbel noise $g_t\in \mathbb{R}^{|V|}$ to the {\color{black} log likelihood} produced in the first forward pass, i.e., $s_t=\log p_{\theta}(\cdot|y_{<t},x)+g_t $, before taking $\argmax$.  
All the elements in $g_t$ are i.i.d. samples drawn from $\texttt{Gumbel}(0,1)$: $g_{t,i}=-\log (-\log(u)), u\sim \texttt{Uniform}(0,1)$ \citep{gumbel1954statistical, DBLP:conf/iclr/JangGP17}.
The $\argmax$ results of $\log p_{\theta}(\cdot|y_{<t},x)$ and $s_t$ may be different and thus enable the decoder to observe a wider variety of input combinations $y^{\text{mix}}$}.

In this paper, we only study mixed CE in teacher forcing and scheduled sampling training under the state-of-the-art Transformer framework due to its wide {\color{black}adoption} and empirical success \citep{NIPS2017_7181}.

\section{Approach}

In this section, we start with the formulation of mixed cross entropy (Mixed CE) and argue its superiority over CE in teacher forcing training and scheduled sampling training. 

\subsection{Mixed Cross Entropy (Mixed CE)}

Given a training instance $(x,y)$, in teacher forcing, mixed CE can be written as:
\begin{equation}
\label{eq2}
\begin{aligned}
        \mathcal{L}_{\textrm{mix}} = - & \big[ (1-\alpha_i) \cdot \sum_{t=1}^n\log p_{\theta}(y_t|y_{<t},x) \\ & + \alpha_i \cdot \sum_{t=1}^n  \log p_{\theta}(\hat{y}_t|y_{<t},x) \big]
\end{aligned}
\end{equation}
{where $\hat{y}_t = \argmax_{1\leq k \leq |V|} \log p_{\theta}({w}_k\vert y_{<t},x)$ \footnote{Here with slight abuse of notation, $y_t, \hat{y}_t$ can be used to denote the token or the token's index in the vocabulary.}}
% where $\hat{y}_t = \argmax_{w \in V} \log p_{\theta}({w} \vert y_{<t},x)$.}

In scheduled sampling, mixed CE is a bit different since the model needs to forward twice (see Section \ref{intro}):
\begin{equation}
\begin{aligned}
    \mathcal{L}_{\textrm{mix}} = - & \big[ (1-\alpha_i)\cdot \sum_{t=1}^n \log p_{\theta}(y_t|y^{\textrm{mix}}_{<t},x) \\ 
    & + \alpha_i \cdot \sum_{t=1}^n \log p_{\theta}(\hat{y}_t|y^{\textrm{mix}}_{<t}, x)
    \big].
\end{aligned}
\label{eq4}
\end{equation}

Here, $\hat{y}_t$ is still $\argmax_{1\leq k \leq |V|} \log p_{\theta}(w_k\vert y_{<t},x)$ obtained in the first forward pass.
$y^{\textrm{mix}}$ {consists of gold tokens in $y$ and the greedily-generated tokens in $\hat{y}$}, as discussed in the previous section.
$\alpha_i$ is a scalar that is related to the $i$-th training iteration, which can be computed as follows:
\begin{equation}
\label{eq3}
    \alpha_i = m \cdot \frac{i}{\textrm{total\_iter}}, 1\leq i \leq \textrm{total\_iter}
\end{equation}
where ``$\textrm{total\_iter}$'' denotes the total training iterations and $m=0.5$ such that the max value of $\alpha_i$ is limited to $0.5$.

{The first part of Eq. \ref{eq2} and \ref{eq4} is the standard CE in both training methods, while the second part is model-dependent CE.}
When $\hat{y}_t=y_t$, mixed CE degenerates to CE, and when $\hat{y}_t \neq y_t$, we justify mixed CE in teacher forcing and scheduled sampling separately.
Specifically, in teacher forcing, we relate machine translation to a noisy label problem, {while in scheduled sampling, we argue mixed CE can make the model less sensitive to input variations, by better simulating test-time behaviours.}

\subsection{Mixed CE in Teacher Forcing: \\Machine Translation as a Noisy Label Problem}\label{noisy}

Noisy labels are unreliable labels that are corrupted from the ground truth \citep{song2020learning}. 
In a one-hot representation, a noisy label places ``1'' on a different index from the ground truth.
This problem often occurs in classification tasks due to untrue annotations at the data collection stage.
In Section \ref{motiv}, we have argued that in CE, $p_{\theta}(\cdot|y_{<t}, x)$ approximating an inappropriate one-hot encoding instead of the underlying ground truth $p^*(\cdot|y_{<t},x)$ might impede generalization.
Ideally, the (soft) label of each token in a sentence should be provided by $p^*(\cdot|y_{<t},x)$ which is context-dependent and the probability mass is scattered over different tokens, especially synonyms.
In that sense, all the $y_t$ (one-hot encoding) in the training set can be deemed as ``\emph{noisy}'' labels.
Thus, we should not fully trust such {\color{black}noisy} labels. 
% {\color{red}Note that this notion of ``noisy'' is more general than in pure classification tasks because the ground truth label $p^*(\cdot|y_{<t},x)$ does not have all of its probability mass concentrated on a single index.}
On the other hand, the empirical success of using {\color{black}noisy} labels as ground truth \citep{DBLP:journals/corr/BahdanauCB14,NIPS2017_7181} suggests that the learned model still preserves information about $p^*(\cdot|y_{<t},x)$.
{That is to say, $p_{\theta}(\cdot|y_{<t}, x)$ is closer to $p^*(\cdot|y_{<t},x)$ than the one-hot encoding of $y_t$.
%So it is reasonable to consider both the one-hot encoding and $p_{\theta}(\cdot|y_{<t}, x)$ as our label.}
So it is reasonable to exploit $p_{\theta}(\cdot|y_{<t}, x)$ during training.
But how much information in $p_{\theta}(\cdot|y_{<t}, x)$ can be used?}

{Previously, we assume that if the model is well-trained and $\hat{y}_t\neq y_t$, $\hat{y}_t$ is very likely to be a synonym of $y_t$.
Under this assumption, $\hat{y}_t$ shall be the useful information provided by $p_{\theta}(\cdot|y_{<t}, x)$ that can be exploited.}
%{\color{red}In the example of Section \ref{motiv}, since ``\emph{decrease}''/``\emph{drop}'' are similar to ``\emph{decline}'' within that context, the model is very likely to allocate more probability mass to such words.}
Therefore, mixed CE also chooses to maximize  $\hat{y}_t$'s probability with a dynamic weight $\alpha_i$ besides the gold target $y_t$.
% Thus, even if the model wrongly predicts ``decrease'' as the most probable token during training, we can still safely maximizes the likelihood of ``decrease'' and this allows for better generalization.
It is worth noting that in Eq. \ref{eq3}, $1-\alpha_i \geq \alpha_i$.
This is because we want our model to be trained on gold data more in the early stage when the model is not yet informative. 
As training progresses, we gradually shift more weights to the model's predictions and finally we treat the gold target $y_t$ and the model's predictions $\hat{y}_t$ equally.

The above learning process essentially assumes there exists a collection of plausible target translations assigned to each input sentence to be translated, where each such target translation serve as a ``label''.
This is essentially a learning problem involving soft or noisy labels.
This label set collapse to a trivial set with only one element, which is the gold target sentence, in the case of standard CE.

% The formulation of mixed CE is similar to Bootstrapping \citep{43273} which is used in object recognition and detection tasks with noisy labels.
% However, there are several differences: 1) mixed CE is used in machine translation with clean labels; 2) when we treat those clean texts as ``noisy'' ones, the ratio of ``noisy'' labels is nearly 100\% while in \citet{43273} this number is much smaller; 3) mixed CE assigns linearly decay coefficients to the two logits with human priors while Bootstrapping selects a fixed value via cross-validation.
Our mixed CE formulation may be reminiscent of other approaches that also adopt soft labels, such as the label smoothing approach \citep{44903}.
Label smoothing assigns $1-\gamma$ ($0<\gamma<1$) probability mass to the gold token, with the remaining $\gamma$ uniformly distributed among all the tokens in the vocabulary, {i.e., $y_t^{LS}=(1-\gamma)\cdot y_t + \frac{\gamma}{|V|}\cdot u$, where $y_t$ is the one-hot encoding and $u$ is a discrete uniform distribution over the whole vocabulary.}
In Section \ref{exp}, we empirically show that the models trained with mixed CE and label smoothing have totally different impacts on the output probability distribution.

\subsection{Mixed CE in Scheduled Sampling: \\Better Simulation of Test-time Behaviours}
\label{ss_mix}

% In scheduled sampling, mixed CE is a bit different since the model needs to forward twice (see Section \ref{intro}):
% \begin{equation}
% \begin{aligned}
%     \mathcal{L}_{\textrm{mix}} = - & \big[ (1-\alpha_i)\cdot \sum_{t=1}^n \log p_{\theta}(y_t|y^{\textrm{mix}}_{<t},x) \\ 
%     & + \alpha_i \cdot \sum_{t=1}^n \log p_{\theta}(\hat{y}_t|y^{\textrm{mix}}_{<t}, x)
%     \big].
% \end{aligned}
% \label{eq4}
% \end{equation}

Now let us focus on the learning objective for mixed CE in the presence of scheduled sampling.
In Eq. \ref{eq4}, the first part of mixed CE is the objective commonly used in scheduled sampling (i.e., Eq. \ref{eq0}), which encourages the model to map the mixed input $y^{\text{mix}}$ to the gold target $y$.
The motivation underlying this is to encourage the model to generate a good output sequence even with its own predictions as input, which is approximated with $y^{\text{mix}}$.
This is essentially a process that is simulating the test-time inputs.
%by mixing the $y$ (only observable during training) and $\hat{y}$ (the only input to the model during testing).

{The second part of the mixed CE objective, however, encourages the output conditioned on $y^{\text{mix}}$ to approximate the greedily-generated sequence $\hat{y}$.
How do we understand such an objective?
Note that from Fig. \ref{sst}, we can see that $\hat{y}$ is in turn the output from the first decoder, which is also parameterized by $\theta$.
In other words, $\hat{y}_{t}=\argmax_{1\leq k \leq |V|} \log p_{\theta}(w_k\vert y_{<t},x)$.
Putting things together, this means with this objective, we are essentially encouraging the model, parameterized by $\theta$, to produce the same output ($\hat{y}$)
regardless of whether the input is the gold input $y$ or the mixed input $y^{\text{mix}}$.
In other words, we are not simply interested in simulating the test-time inputs now, but we would also like to make $y^{\text{mix}}$ and $y$ indistinguishable when serving as inputs to the model.
This effectively requires the model to share the same internal {\em behaviour} (e.g., similar internal neural states) when the input is either $y$ or $y^{\text{mix}}$, while the former is related to the training phase, and the latter is related to the testing phase.

Overall, our mixed CE approach is essentially simulating the test-time behaviour (with the second term), while encouraging the model to learn to predict the gold output $y$ with the simulated test-time input $y^{\text{mix}}$ (with the first {term}).
}

\section{Experiments}
\label{exp}

In this section, we conducted experiments to verify the effectiveness of mixed CE in teacher forcing and scheduled sampling on several benchmark datasets with different sizes, WMT'16 Romanian-English (Ro-En, 610K pairs), WMT'16 Russian-English (Ru-En, 2.1M pairs) and WMT'14 English-German (En-De, 4.5M pairs). 
We begin with some training details of all experiments and then we study mixed CE in teacher forcing and scheduled sampling separately.

\subsection{Experimental Setup}
\label{exp_setup}

We used the preprocessed WMT'16 Ro-En dataset from \citet{lee2018deterministic} with vocabulary sizes of Romanian and English being 27,591 and 21,175, respectively.
For WMT'14 En-De, we used the script from Fairseq \citep{ott2019fairseq}\footnote{\url{https://github.com/pytorch/fairseq}} for preprocessing  (we used newstest2013 as the validation set instead following \citet{zhang-etal-2019-bridging}).
The vocabulary size of English is 40,511 while for German it is 42,735.
We used a script similar to WMT'14 En-De to preprocess WMT'16 Ru-En but separate BPE codes \citep{sennrich-etal-2016-neural} for Russian and English with 24K merge operations, resulting in 26,327 and 26,319 tokens in each vocabulary were adopted.
We used separate token embeddings for source and target languages in all 3 datasets.
Standard base Transformer \citep{NIPS2017_7181} architecture was used in the experiment. 
We trained the model for totally 8,000/45,000/80,000 iterations for Ro-En/Ru-En/De-En datasets with each batch containing 12,288$\times$4/12,288$\times$4/12,288$\times$8 tokens.
All models were pre-trained with CE for 5 epochs.
We used the Adam \citep{DBLP:journals/corr/KingmaB14} optimizer with $\beta_1=0.9, \beta_2=0.98$.
Learning rate is $0.0007$ and will be reduced by half if the BLEU \citep{papineni-etal-2002-bleu} score on validation set does not increase in the last 4 epochs.
Unless otherwise specified, we also used label smoothing ($\gamma=0.1$) in our experiments.
The decay strategy for scheduled sampling (see Section \ref{intro}) that we use follows \citet{leblond2018searnn}:
\begin{equation}
    \epsilon_i = d^{i/{\textrm{total\_iter}}}, 0 < d < 1, 1\leq i \leq \textrm{total\_iter}
\end{equation}
where $\epsilon_i$ is related to $i$-th training iteration and decays as training proceeds.
% $i$ denotes $i$-th training iteration and “total\_iter” denotes the total number of training iterations.
For Ro-En, $d=0.7$ while for Ru-En/En-De, $d=0.8$.
We saved a checkpoint after training the model for each epoch and we selected the best checkpoint based on the performance on the validation set, which we refer to as ``Single''.
Besides, we also reported the performance of the average models obtained by averaging either the last 5 checkpoints or top 5 checkpoints, depending on the performance on the validation set.
We refer to this average model as ``Average''.
% We report the results of the average model according to BLEU score on validation set since different training settings often lead to different convergence speed.

\subsection{Mixed CE in Teacher Forcing}
\label{mix_ce_in_tf}
We use a base Transformer trained with CE as a baseline and compare it with mixed CE.
We also compare mixed CE with a loss function that is originally designed for neural machine translation, Dual Skew Divergence (DSD) loss \citep{xiao2019dual}, {which minimizes the forward and reverse $\alpha$-skew KL divergence between empirical data distribution and model prediction}. 
% \footnote{We also implemented Minimum Risk Training \citep{shen-etal-2016-minimum} following \citet{edunov-etal-2018-classical}. But the training is not stable and extremely time-consuming (20$\times$ slower with beam size 8. The suggested beam size is 100 in \citet{shen-etal-2016-minimum} and 16 in \citet{edunov-etal-2018-classical}.) {and thus we failed to find a good set hyper-parameters on the 3 data sets, even with the help from the authors of \citet{edunov-etal-2018-classical}.}}.
The best-performing hyperparameters in the original DSD paper were used in our experiments.
Besides, according to \citet{xiao2019dual}, DSD only works after the model has been pre-trained with CE, otherwise the performance would drop quickly.
The number of pre-training iterations was chosen based on the performance on the validation set.
\textcolor{black}{Moreover, we also compare mixed CE with a sequence level self-distillation method (self-dist) \citep{kim-rush-2016-sequence, pmlr-v80-furlanello18a}.\footnote{There is another approach is also called self-knowledge distillation \textcolor{black}{\citep{hahn-choi-2019-self} where the $\alpha$ is computed from the scaled Euclidean distance between the embeddings of the gold token and the model's prediction. We tried several scale factors but failed to find one that is able to outperform the baseline in our training settings. We discuss this in the appendix.}}
We first train a Transformer with CE and re-generate the target {\color{black}sequence} $y$ in the training data using greedy search (we choose greedy search because it corresponds to the $\argmax$ operation in Eq. \ref{eq2}), obtaining new $\hat{y}$.
Note that during greedy search, we force the length of $\hat{y}$ to be the same as $y$.
Then we use the same loss function as Eq. \ref{eq2} but replace {\color{black}the original} $\hat{y}$ with our distilled $\hat{y}$ to train the model. }
Results are shown in Table \ref{table1}.
All results are averaged over 3 runs.
\textcolor{black}{{\color{black}Self-dist} does not perform well and we conjecture this is because the model has difficulty fitting  two different modes at the same time.}
Mixed CE outperforms CE in most settings even though the performance gap is smaller on larger data sets.
Compared to DSD, mixed CE generally gives better performance and mixed CE is much easier to train.
It can be observed that mixed CE typically brings larger improvement in single model testing.
In Ru-En, the single model trained with mixed CE even approaches the average model trained with CE.
Besides, the improvements on average models seem to be more significant when we apply greedy decoding (beam size = 1).
All the test sets used so far are single-reference sets, which may not demonstrate mixed CE's ability to exploit synonyms.
So we further experiment with a multi-reference set as well as a paraphrased single-reference set.

% \begin{table}[t]
% \caption{BLEU scores on test sets of Transformers trained with CE and mixed CE. The results of beam search decoding with beam size 1/5 are presented. All results are averaged over 2 runs. }
% \label{table1}
% \vskip 0.15in
% \begin{center}
% \begin{small}
% \scalebox{1}
% {
% \begin{sc}
% \begin{tabular}{ccccc}
% \toprule
% Data set & Loss & Single & Average  \\
% \midrule
%  \multirow{3}{*}{ Ro-En}   & CE & 30.43/31.28 & 31.84/32.36 \\
%  & DSD & \textbf{31.37}/31.91 & 31.74/32.51 \\
%  & Mixed CE & 31.24/\textbf{32.11} & \textbf{32.58/33.15} \\
% % & & 0.0364/0.0043 & 0.0599/0.0933 \\
%  \midrule

%  \multirow{3}{*}{ Ru-En} & CE & 28.88/30.14 & 29.51/30.82 \\
%     & DSD & 28.91/30.23 & 29.69/30.85\\
%     & Mixed CE & \textbf{29.54/30.68} & \textbf{30.11/31.01}  \\
%  %   & & 0.007/0.0096 & 0.0042/0.0719\\
%  \midrule
%  \multirow{3}{*}{ En-De} & CE & \textbf{26.38}/27.04 & 26.61/27.47 \\
%       & DSD & 26.30/26.93 & 26.60/27.22 \\
%       & Mixed CE & 26.37/\textbf{27.30} & \textbf{26.74/27.53} \\
% %      & &    & \\
% \bottomrule
% \end{tabular}
% \end{sc}
% }
% \end{small}
% \end{center}
% \vskip -0.1in
% \end{table}

\begin{table}[t]
\caption{BLEU scores on test sets of Transformers trained with CE and mixed CE. The results of beam search decoding with beam size 1/5 are presented. \textcolor{black}{All results are averaged over 3 runs}. }
\label{table1}
%\vskip 0.15in
\begin{center}
\begin{small}
\scalebox{1}
{
\begin{sc}
\begin{tabular}{ccccc}
\toprule
Data set & Loss & Single & Average  \\
\midrule
 \multirow{4}{*}{ Ro-En}   & CE & 30.63/31.42 & 32.07/32.59 \\
 & DSD & \textbf{31.17}/31.80 & 32.03/32.74 \\
 & Self-Dist &  28.65/31.45  & 31.66/32.61 \\
 & Mixed CE & \textbf{31.17}/\textbf{32.02} & \textbf{32.63/33.25} \\
% & & 0.0364/0.0043 & 0.0599/0.0933 \\
 \midrule

 \multirow{4}{*}{ Ru-En} & CE & 28.87/30.24 & 29.48/30.79 \\
    & DSD & 28.89/30.30 & 29.69/30.90\\
     & Self-Dist & 28.76/30.34 & 29.32/30.63 \\
    & Mixed CE & \textbf{29.59/30.74} & \textbf{30.14/31.05}  \\
 \midrule
 \multirow{4}{*}{ En-De} & CE & 26.23/26.91 & 26.67/27.41 \\
       & DSD & 26.10/26.84 & 26.66/27.30 \\
        & Self-Dist & 24.15/25.98  &  24.23/25.91 \\
      & Mixed CE & \textbf{26.32}/\textbf{27.28} & \textbf{26.72/27.61} \\
\bottomrule
\end{tabular}
\end{sc}
}
\end{small}
\end{center}
%\vskip -0.1in
\end{table}

\paragraph{Additional Reference}
We also conducted experiments to see how CE-based, mixed CE-based models perform when we use a new set of references.
We chose two additional reference sets: 1) a multiple-reference test set of WMT'14 En-De \citep{ott2018analyzing}\footnote{\url{https://github.com/facebookresearch/analyzing-uncertainty-nmt}} where there are 10 additional references (the original reference is excluded) for each of the 500 test sentences taken from the original test set; 2) a paraphrased as-much-as-possible version of the original WMT'19 En-De reference \cite{freitag-bleu-paraphrase-references-2020}\footnote{\url{https://github.com/google/wmt19-paraphrased-references}}.
In the first set, 10 human reference translations of the same source sentence cover a broader reference space and exhibit certain amount of diversity in lexical choices.
This will help us verify the effectiveness of mixed CE since it is expected to maximize the probability of synonyms of training tokens.
Nevertheless, these human translations are often influenced by source sentences and thus tend to have a monotonic alignment to the source side and a relatively simple vocabulary \citep{koppel-ordan-2011-translationese, freitag-bleu-paraphrase-references-2020}.
Therefore, we further selected the second set in which each reference translation is paraphrased from the original reference by human experts and differs significantly from the original one in word choices (more advanced synonyms) and sentence structures (non-monotonic alignment) but keeps the same  meaning.
According to \citet{freitag-bleu-paraphrase-references-2020}, BLEU scores on this paraphrased test set correlates better with human ratings than the original reference and hence we should be more confident about the superiority of mixed CE if it yields higher BLEU.
Note that when evaluated on the WMT'19 En-De test set, the {\color{black}models were} still trained with the WMT'14 training data.

% \begin{table}[t]
% \caption{BLEU improvement of mixed CE over CE on 10 additional references of WMT'14 En-De test set. All results are averaged over 2 runs.}
% \label{add_ref}
% \vskip 0.15in
% \begin{center}
% \begin{sc}
% \resizebox{0.45\textwidth}{!}{
% \begin{tabular}{lcccc}
% \toprule
%  \multirow{2}{*}{ REF} & \multicolumn{2}{c}{AVG} & \multicolumn{2}{c}{TOP}   \\
%  & CE & Mixed CE & CE & Mixed CE \\
% \hline
%   ref 1 & 36.71 & \textbf{37.43 (+0.72)} & 38.64 & \textbf{39.27 (+0.63)} \\
%  ref 2 & 47.26 & \textbf{48.70 (+1.43)}  & 50.09 &  \textbf{51.73 (+1.64)} \\
%  ref 3 & 42.51 & \textbf{43.47 (+0.96)}  & 44.77 & \textbf{46.14 (+1.37)} \\
%  ref 4 & 29.08 & \textbf{29.88 (+0.80)} & 30.48 &  \textbf{31.14 (+0.66)} \\
%  ref 5 & 31.46 & \textbf{32.65 (+1.19)} & 33.13 & \textbf{34.31 (+1.18)} \\
%  ref 6 & 26.26 & \textbf{26.88 (+0.62)} & 27.36 & \textbf{27.96 (+0.60)} \\
%  ref 7 & 42.24 & \textbf{43.06 (+0.82)} & 44.60 & \textbf{45.08 (+0.48)} \\
%  ref 8 & 32.29 & \textbf{33.21 (+0.92)} & 33.65 & \textbf{34.64 (+0.99)} \\
%  ref 9 & 28.45 & \textbf{29.11 (+0.66)} & 29.69 & \textbf{30.30 (+0.61)} \\
%  ref 10 & 33.63 & \textbf{33.97 (+0.34)} & 35.31 & \textbf{35.71 (+0.40)} \\
%  \midrule
%  Mean & 34.99 & \textbf{35.84 (+0.85)} & 36.77 & \textbf{37.62 (+0.85)} \\

% \bottomrule
% \end{tabular}}
% \end{sc}
% \end{center}
% \vskip -0.1in
% \end{table}

\begin{table}[t]
\caption{\textcolor{black}{}BLEU improvement of mixed CE over CE on 10 additional references of WMT'14 En-De test set. \textcolor{black}{All results are averaged over 3 runs}.}
\label{add_ref}
%\vskip 0.15in
\begin{center}
\begin{sc}
\resizebox{0.45\textwidth}{!}{
\begin{tabular}{lcccc}
\toprule
 \multirow{2}{*}{ REF} & \multicolumn{2}{c}{AVG} & \multicolumn{2}{c}{TOP}   \\
 & CE & Mixed CE & CE & Mixed CE \\
\hline
  ref 1 & 36.73 & \textbf{37.32 (+0.59)} & 38.61 & \textbf{39.13 (+0.52)} \\
 ref 2 & 47.48 & \textbf{48.50 (+1.02)}  & 50.08 &  \textbf{51.36 (+1.28)} \\
 ref 3 & 42.59 & \textbf{43.25 (+0.66)}  & 44.89 & \textbf{45.89 (+1.00)} \\
 ref 4 & 28.93 & \textbf{29.78 (+0.85)} & 30.29 &  \textbf{30.98 (+0.69)} \\
 ref 5 & 31.75 & \textbf{32.53 (+0.78)} & 33.48 & \textbf{34.18 (+0.70)} \\
 ref 6 & 26.41 & \textbf{26.83 (+0.42)} & 27.60 & \textbf{27.96 (+0.36)} \\
 ref 7 & 42.18 & \textbf{42.89 (+0.71)} & 44.37 & \textbf{44.90 (+0.53)} \\
 ref 8 & 32.36 & \textbf{33.05 (+0.69)} & 33.77 & \textbf{34.55 (+0.78)} \\
 ref 9 & 28.51 & \textbf{29.03 (+0.52)} & 29.65 & \textbf{30.27 (+0.62)} \\
 ref 10 & 33.75 & \textbf{33.94 (+0.19)} & 35.23 & \textbf{35.68 (+0.45)} \\
 \midrule
 Mean & 35.07 & \textbf{35.71 (+0.64)} & 36.80 & \textbf{37.49 (+0.69)} \\

\bottomrule
\end{tabular}}
\end{sc}
\end{center}
%\vskip -0.1in
\end{table}

We listed the BLEU score improvement of the model trained with mixed CE over the model with CE on 10 additional references in Table \ref{add_ref}.
We used beam search (beam size 10) to generate 10 hypotheses for each source sentence.
We reported the average (AVG) and the largest (TOP) BLEU scores of the 10 hypotheses with respect to each reference.
Mixed CE consistently outperforms CE across all additional references with the average improvement being {\color{black}0.64/0.69} BLEU. %LW: shouldn't this be 0.64/0.69?}
This suggests that mixed CE can assist in generating hypotheses that align better with human translations in a general sense.

The results for the second reference set is shown in Table \ref{paraphrase}.
We used beam search (beam size 1/10) and sampling (sampled 100 times and selected the sentence with the highest average {\color{black} log likelihood)}
%{\color{blue}LW: how did you exactly do the sampling?}
to generate hypotheses.
As a reference, the BLEU score of the machine translation system trained with WMT'19 En-De training data (it has much more sentence pairs than WMT'14 En-De, 38.8M vs 4.5M) on this paraphrased test set is 12.5 \citep{freitag-bleu-paraphrase-references-2020}.
In addition, it is worth to note that BLEU score on this reference set is much lower than on the original one due to less $n$-gram matches.
Despite a significant change in reference, mixed CE still surpasses CE (+0.34/+0.27/+1.01 BLEU), especially when using sampling decoding.
% For the sake of comparison, \citet{freitag-bleu-paraphrase-references-2020} reported that the BLEU score improvement of the machine translation system augmented with Automatic-Post-Editing/Back-Translation \citep{DBLP:conf/wmt/FreitagCR19, sennrich-etal-2016-improving} on this paraphrased set is 0.2/0.4 BLEU.
These results on the two additional references again confirm that mixed CE exploits useful information in the model predictions.

\begin{figure}
    \centering
    \includegraphics[scale=0.5]{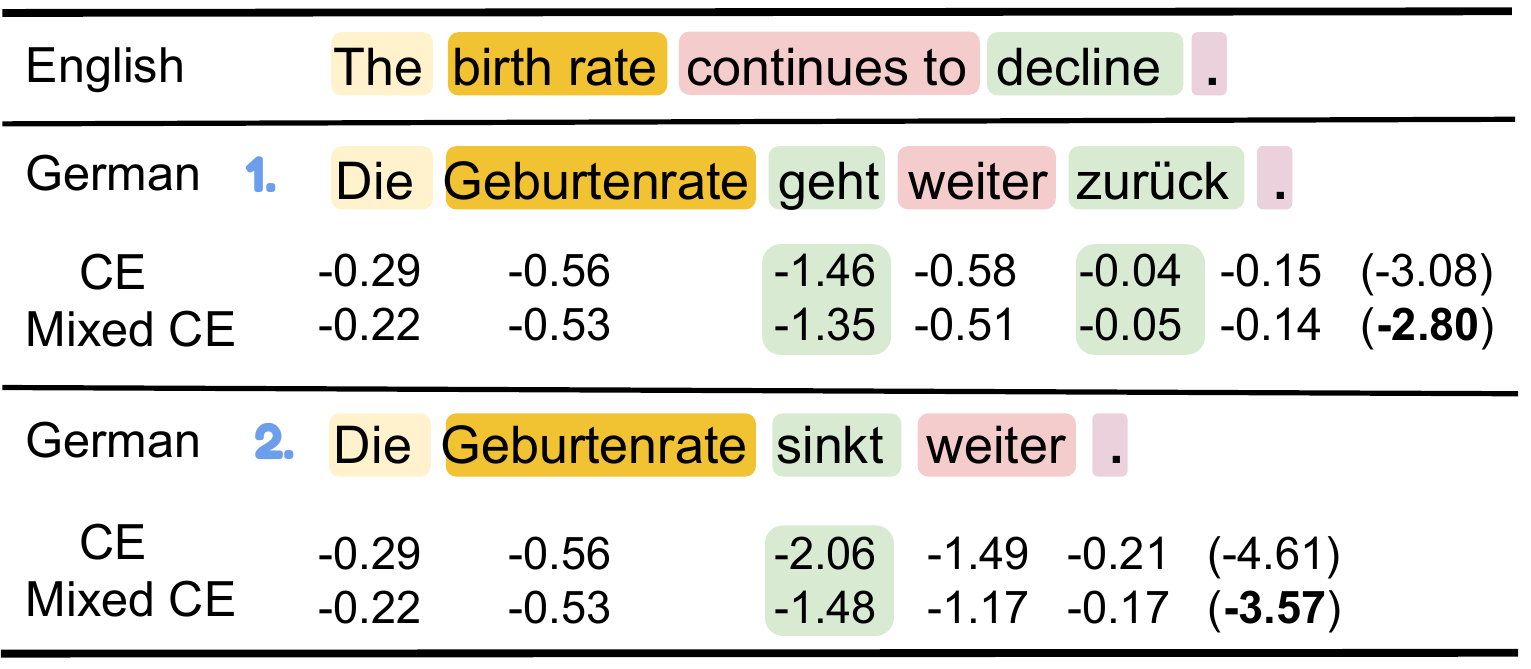}
    \caption{Multiple translations with their positional scores. We used beam search to translate the English sentence into German and we selected the top-2 candidates. Words in the same color have the same meaning.}
    \label{trans}
\end{figure}

Last, we give a concrete example of English-German translation produced by CE-trained and mixed CE-trained models in Fig. \ref{trans}. 
We plotted each token's $\log_2$ score given by the models.
For the synonyms ``{\em sinkt}'' and ``{\em geht zurück}'' (both mean ``{\em decline}''), CE-trained model gives them scores $-1.5$ and $-2.06$ while mixed CE-trained model produces $-1.4$ and $-1.48$ respectively.
The above example shows that: 1) {\color{black}CE-based models contain useful synonym information since they can find synonyms in the top-2 candidates through beam search}; 2) { mixed CE-based models can allocate more probability {\color{black}mass} to those synonyms}.
It can also be observed that mixed CE-based model {\color{black}concentrates more {\color{black}(in terms of probability mass)} on the top-2 candidate translations than CE-based model {\color{black}(-2.8 vs -3.08 ; -3.57 vs -4.61)}, even though both models produce the same candidates.}

\begin{table}[t]
\caption{BLEU scores of beam search/sampling results on WMT'19 En-De paraphrased test set. As a reference, \citet{freitag-bleu-paraphrase-references-2020} reported that the BLEU score improvement of the machine translation system augmented with Automatic-Post-Editing/Back-Translation \citep{DBLP:conf/wmt/FreitagCR19, sennrich-etal-2016-improving} on this paraphrased set {\color{black}was} 0.2/0.4 BLEU.}
\label{paraphrase}
%\vskip 0.15in
\begin{center}
\scalebox{0.9}{
\begin{sc}
\begin{tabular}{cccc}
\toprule
  Loss & Beam 1 & Beam 10 & Sampling    \\
\midrule
  CE & 11.26 & 11.67 & 8.89 \\
Mixed CE & \textbf{11.60} & \textbf{11.94} & \textbf{9.90} \\

\bottomrule
\end{tabular}
\end{sc}
}
\end{center}
%\vskip -0.1in
\end{table}

\paragraph{Comparison with Label Smoothing}
Since label smoothing and mixed CE have a similar formulation  (see Section \ref{noisy}), we also studied the different impacts these two techniques have on the model.
Label smoothing improves generalization by penalizing confident predictions \citep{pereyra2017regularizing, NEURIPS2019_f1748d6b}.
To figure out whether mixed CE works in the same way, we first trained 4 different models on WMT'14 En-De: 1) without label smoothing and mixed CE; 2) only with label smoothing; 3) only with mixed CE; 4) with both label smoothing and mixed CE.
Then each model generated 5 hypotheses for each sentence in the validation set using sampling decoding and we calculated the Pairwise-BLEU (PB) \citep{shen2019mixture} among these hypotheses.
PB is used to measure the diversity of the generated hypotheses.
The more diverse the hypotheses are, the lower the PB is.
Furthermore, a flat probability distribution tends to generate more diverse hypotheses if we use sampling decoding and thus we can measure the sharpness of the output distribution with PB.
{We also computed the BLEU score of the 4 models on the validation set using beam search (beam size 5).}

The results are shown in Table \ref{pairwise}.
We can see that label smoothing leads to lower PB than the baseline (the first row), indicating a more flat distribution.
Mixed CE, however, gives much higher PB which suggests a more peaked distribution.
When combining label smoothing and mixed CE, the resulting PB lies in the middle ground.
Thus, mixed CE has totally different impacts on the output distribution {\color{black} as can be measured from PB}.
{{\color{black}Based on} the BLEU score, we can see that both label smoothing and mixed CE boost model performance and the combination of them yields even better results.
{ 
These facts show the two approaches work differently (according to PB) and {\color{black}are} able to complement each other (according to BLEU).}
}

\begin{figure}
    \centering
    \includegraphics[scale=0.5]{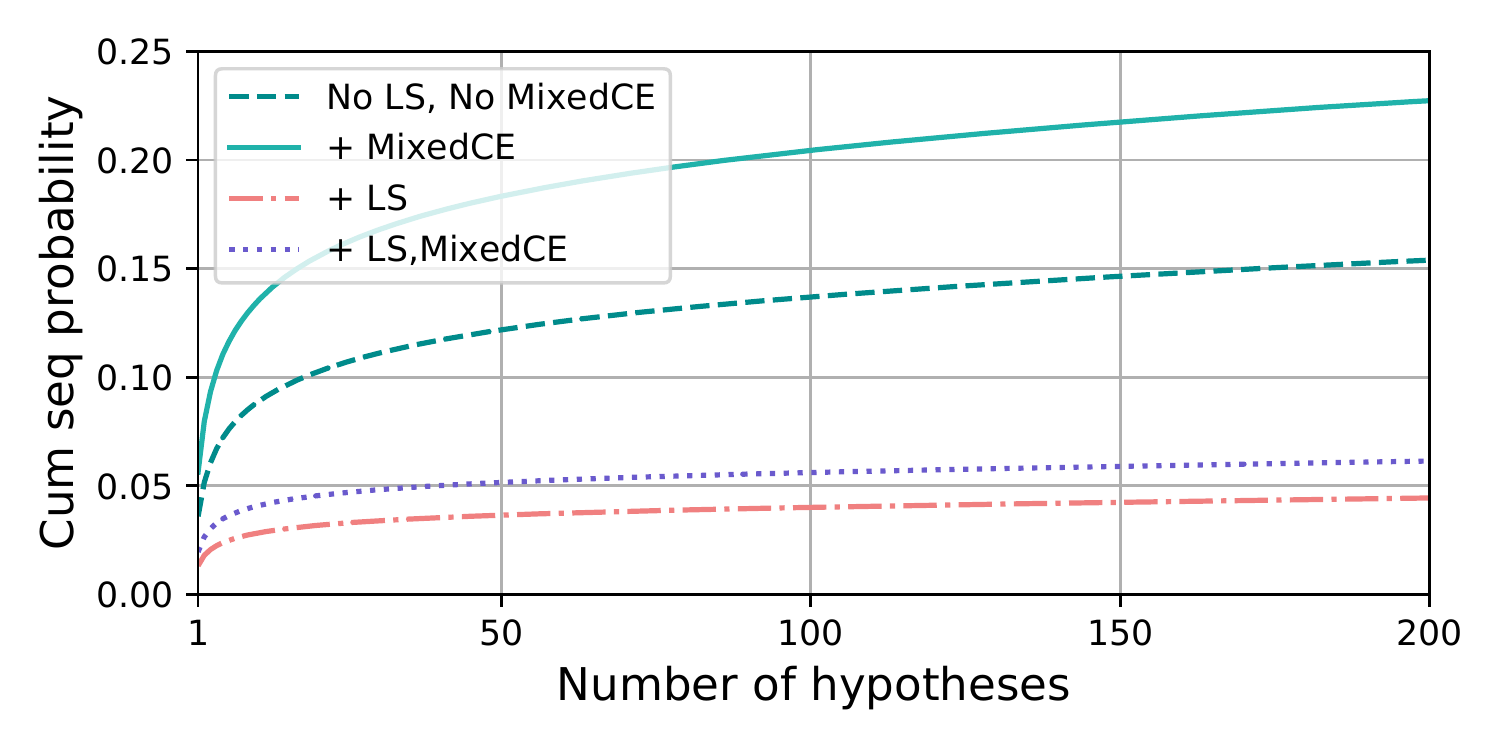}
    \caption{Cumulative sequence probability of generated hypotheses using beam search with beam size 200 on WMT'14 En-De validation set. }
    \label{cum_seq_prob}
    \vskip -0.1in
\end{figure}

To study the distribution properties of 4 models quantitatively, we calculated the cumulative sequence probability\footnote{The sum of the probabilities of the generated hypotheses.} of all the hypotheses generated by beam search \citep{ott2018analyzing}.
The results are shown in Fig. \ref{cum_seq_prob}.
We can see that label smoothing smears the probability mass evenly in the whole hypothesis space according to the linear increasing trends in cumulative probability.
Mixed CE tends to assign the probability mass to the top-scoring candidates as revealed by the sharp increase {of the cumulative probability} in the top-50 candidates.
{This is also consistent with the findings in Fig. \ref{trans}.}
{\color{black}As we increase the number of hypotheses} to $>50$, the cumulative sequence probability increases linearly (approximately).

All the evidence above have proved that mixed CE is different from label smoothing despite a similar formulation.
For generation tasks, it is desirable to have more probability mass assigned to { tokens that are relevant to the context only, while for irrelevant tokens their probabilities of appearing in a specific context shall be very low, if  at all possible (e.g., the word ``{\em chicken}'' is likely an irrelevant word with respect to the text discussing aircraft maintenance)}.
On the other hand, it is also necessary to avoid overfitting by penalizing confident predictions \citep{pereyra2017regularizing, NEURIPS2019_f1748d6b}.
Therefore, using both label smoothing and mixed CE together may allow us to have the best of both worlds, arriving at the optimal results in practice.

\begin{table}[t]
\caption{PB, BLEU on WMT'14 En-De validation set. Pairwise-BLEU is obtained using sampling decoding while the BLEU score is obtained using beam search. LS is short for label smoothing.}
\label{pairwise}
%\vskip 0.15in
\begin{center}
\scalebox{0.9}
{
\begin{sc}
\begin{tabular}{ccc}
\toprule
  Loss & PB {\color{black}($\downarrow$)} & BLEU {\color{black}($\uparrow$)}   \\
\midrule
   No LS, No Mixed CE & 17.52 & 25.81  \\
+ LS & {\color{white}0}5.22 & 26.48  \\
+ Mixed CE & 25.99 & 26.26 \\
+ LS, mixed CE & {\color{white}0}7.79 & \textbf{26.75} \\

\bottomrule
\end{tabular}
\end{sc}
}
\end{center}
%\vskip -0.1in
\end{table}

\subsection{Mixed CE in Scheduled Sampling}

In scheduled sampling, we tested mixed CE on two different baselines, standard SS \citep{NIPS2015_5956} and \emph{word oracle} \citep{zhang-etal-2019-bridging}.\footnote{\citet{zhang-etal-2019-bridging} compared their approach with other reinforcement learning based baselines and thus we did not include those comparisons here.}
% Different from standard SS, Word Oracle adds some Gumbel noise $g_t\in \mathbb{R}^{|V|}$ to the logits $\log p_{\theta}(\cdot|y_{<t},x) \in \mathbb{R}^{|V|}$ produced in the first forward pass (see Fig. \ref{sst}) before taking $\argmax$: $s_t=\log p_{\theta}(\cdot|y_{<t},x)+g_t $.
% All the elements in $g_t$ are i.i.d. samples drawn from $\text{Gumbel}(0,1)$: $g_{t,i}=-\log (-\log(u)), u\sim \text{Uniform}(0,1)$.
% The $\argmax$ results of $\log p_{\theta}(\cdot|y_{<t},x)$ and $s_t$ may be different and thus enable the decoder to observe more input combinations $y^{\text{mix}}$ and be more robust to input variations.
Note that in word oracle experiments, the $\hat{y}_t$ in the loss function (see Eq. \ref{eq4}) is still obtained from $\argmax_{1\le k \le |V|} \log p_{\theta}(w_k|y_{<t},x)$ while $y^{\text{mix}}$ in Fig. \ref{sst} is obtained by mixing $y_t$ with $\hat{y}_t=\argmax_{1\le k \le |V|} s_{t,k}$ ($k$-th element of $s_t$; {\color{black}for $s_{t}$, see Sec. \ref{intro}}).
The results are shown in Table \ref{table4}.
We can see that mixed CE surpasses CE on small and medium data sets by a large margin but this gap is smaller on a larger data set.

\begin{table}[t]
\caption{BLEU scores on test sets of Transformers trained with CE and mixed CE. The results of beam search decoding with beam size 1/5 are presented. \textcolor{black}{All results are averaged over 3 runs}.}
\label{table4}
%\vskip 0.15in
\begin{center}
\scalebox{0.85}
{
\begin{sc}
\begin{tabular}{cccc}
\toprule
 \multirow{2}{*}{Data set}& \multirow{2}{*}{Loss} &\multicolumn{2}{c}{Schedueld Sampling} \\
 \cline{3-4}
 & & Single & Average  \\
\midrule
 \multirow{2}{*}{ Ro-En}   & CE & 30.71/31.72 & 32.29/33.05 \\
 & Mixed CE & \textbf{31.71/32.53} & \textbf{32.88/33.45} \\
 \midrule

 \multirow{2}{*}{ Ru-En} & CE & 29.28/30.63 & 29.62/30.83 \\
    & Mixed CE & \textbf{30.19/31.23} & \textbf{30.47/31.39}  \\
 \midrule
 \multirow{2}{*}{ En-De} & CE & 26.36/27.29 & 26.84/27.56 \\
      & Mixed CE & \textbf{26.75/27.57} & \textbf{26.99/27.71} \\ 
\midrule
\midrule

 \multirow{2}{*}{Data set}& \multirow{2}{*}{Loss} &\multicolumn{2}{c}{Word Oracle} \\
 \cline{3-4}
 & & Single & Average  \\
\midrule
 \multirow{2}{*}{ Ro-En}   & CE & 31.71/32.37  & 33.05/33.76  \\
 & Mixed CE & \textbf{32.43/33.06} & \textbf{33.66/34.14} \\
 \midrule

 \multirow{2}{*}{ Ru-En} & CE & 29.40/30.61 & 29.87/31.00 \\
    & Mixed CE & \textbf{30.24/31.09} & \textbf{30.72/31.50}  \\
 \midrule
 \multirow{2}{*}{ En-De} & CE & 26.66/27.45 & \textbf{26.94}/27.71 \\
      & Mixed CE & \textbf{26.81/27.80} & \textbf{26.94/27.88} \\ 
      
\bottomrule
\end{tabular}
\end{sc}
}
\end{center}
%\vskip -0.1in
\end{table}

But the BLEU improvements alone are not enough to demonstrate the necessity of the second part in Eq. \ref{eq4}.
Hence, we further modify the way to calculate $\hat{y}_t$ which is produced in the first pass:
\begin{equation}
\label{newy1}
    \hat{y}_t = \text{Rand}\Big(\text{Top-2}_{1\leq k \leq |V|}\big(\log p_{\theta}(w_k|y_{<t},x)\big)\Big)
\end{equation}
We randomly sampled the highest or the second highest scoring token to substitute the $\hat{y}_t$ in Eq. \ref{eq4}.
The results are shown in Table \ref{comparison}.
We can see that in the third row,  top-2 mixed CE significantly impairs the model's performance.
We surmise that if the model can correctly predict the gold token as the highest-scoring token and yet we still maximize the likelihood of the second highest-scoring token, this will confuse the model and lead to a drop in performance.
So we further modified Eq. \ref{newy1}:
\begin{equation} 
    \hat{y}_t  = \begin{cases}
         y_t  ,  \text{if $y_t = \argmax_{1\leq k \leq |V|} \log p_{\theta}(w_k|y_{<t},x)$} &  \\
         \text{Rand($V$)}, \text{otherwise} & 
    \end{cases}
\end{equation}
$\text{Rand}(V)$ denotes a random token in the vocabulary.
The results can be found in the 4-th row in Table \ref{comparison}.
Random mixed CE is better than top-2 mixed CE but still underperforms mixed CE, indicating the importance of approximating the outputs conditioned on gold inputs.
It is also worth to note that random mixed CE outperforms CE for which we conjecture that random mixed CE provides a form of regularization.
% We also tried a soft version of mixed CE as shown in Eq. \ref{eq8} which is denoted as soft mixed CE.
% Obviously, the second part in Eq. \ref{eq8} prevents the model from learning and thus harms performance.
% \begin{equation}
% \label{eq8}
% \begin{aligned}
%      \mathcal{L}_{\textrm{mix}} = -\sum_{t=1}^n  \big[ (1-\alpha_i)\cdot \log p_{\theta}(y_t|y^{\textrm{mix}}_{<t},x) \\ 
%      + \alpha_i \cdot  \sum_{k=1}^{|V|} q_{\theta}(y_{t,k}|y_{<t},x)\cdot \log p_{\theta}(\hat{y}_{t,k}|y^{\textrm{mix}}_{<t}, x)
%     \big]
% \end{aligned}
% \end{equation}
We also tried a soft version of mixed CE in scheduled sampling as shown in Eq. \ref{eq8}, which is denoted as soft mixed CE.
Here $q_{\theta}(w_{k}|y_{<t},x)$ denotes the model's prediction in the first pass for the $k$-th token in the vocabulary at time step $t$ conditioned on gold prefixes.
 The results can be found in the 5-th row in Table \ref{comparison}.
The results indicate that we do not need to simulate the outputs in the first pass that are considered unlikely by the model in order to homogenize train/test behaviors.
\begin{equation}
\label{eq8}
\begin{aligned}
     \mathcal{L}_{\textrm{mix}} = -\sum_{t=1}^n  \big[ (1-\alpha_i)\cdot \log p_{\theta}(y_t|y^{\textrm{mix}}_{<t},x) \\ 
     + \alpha_i \cdot  \sum_{k=1}^{|V|} q_{\theta}(w_k|y_{<t},x)\cdot \log p_{\theta}(w_{k}|y^{\textrm{mix}}_{<t}, x)
    \big]
\end{aligned}
\end{equation}

\begin{table}[t]
\caption{BLEU scores of Transformers trained with different loss functions on the WMT'16 Ro-En validations sets. }
\label{comparison}
% \vskip 0.15in
\begin{center}
\scalebox{0.9}{
\begin{sc}
\begin{tabular}{lcc}
\toprule
 Loss & SS & Word Oracle  \\
\midrule
   CE & 32.66 & 33.82 \\
  Mixed CE & \textbf{33.64} & \textbf{34.51} \\
  Top-2 Mixed CE & 32.17 & 32.76\\
  Random Mixed CE & 33.26 & 34.18 \\
   Soft Mixed CE & 32.03 & 33.08\\
      
\bottomrule
\end{tabular}
\end{sc}
}
\vspace{-15pt}
\end{center}
% \vskip -0.1in
\end{table}

\subsection{Combining Two Mixed CE}

In scheduled sampling training, we also considered making use of the model's predictions in the second forward pass as what we do in teacher forcing by modifying mixed CE in scheduled sampling as follows:
\begin{equation}
\label{doublemix}
\begin{aligned}
         \mathcal{L}_{\text{mix}} = -\sum_{t=1}^n\big[&(1-\alpha_i)\cdot \log p_{\theta}(y_t|y_{<t}^{\text{mix}},x) \\
         &+ \frac{\alpha_i}{2} \cdot \big( \log p_{\theta}(\hat{y}_t|y_{<t}^{\text{mix}},x) \\
         &+ \log p_{\theta}(\tilde{y}_t|y_{<t}^{\text{mix}},x)  \big)  \big]
\end{aligned}
\end{equation}

Here, $\tilde{y}_t$ is the greedy decision in the second forward pass, i.e., $\tilde{y}_t = \argmax_{1\leq k \leq |V|} \log p_{\theta}(w_k|y_{<t}^{\text{mix}},h)$.
We denote this new loss as double mixed CE.
The results are shown in Table \ref{double}.
\begin{table}[t]
\caption{BLEU scores of Transformers trained with \emph{double mixed CE} and \emph{mixed CE-2nd pass} on validations sets. }
\label{double}
\vskip 0.15in
\begin{center}
\begin{small}
\scalebox{0.9}{
\begin{sc}
\begin{tabular}{lccc}
\toprule
 Loss & Ro-En & Ru-En & En-De  \\
\midrule
   CE & 33.82 & 29.83 & 26.51\\
   Mixed CE & \textbf{34.51} & \textbf{30.46} & 26.88\\
   Double Mixed CE & 34.23 & \textbf{30.46} & \textbf{27.06}\\
   Mixed CE-2nd pass & 33.84 & 30.16 & 26.83\\
      
\bottomrule
\end{tabular}
\end{sc}
}
\vspace{-15pt}
\end{small}
\end{center}

\end{table}
The performance does not vary much on all data sets.
But does this mean $\tilde{y}_t$ and $\hat{y}_t$ are equally informative?
So we further modified Eq. \ref{doublemix}:
\begin{equation}
\begin{aligned}
         \mathcal{L}_{\text{mix}} = -\sum_{t=1}^n\big[&(1-\alpha_i)\cdot \log p_{\theta}(y_t|y_{<t}^{\text{mix}},x) \\
         &+ \alpha_i \cdot \log p_{\theta}(\tilde{y}_t|y_{<t}^{\text{mix}},x)  \big]
\end{aligned}
\end{equation}

We denote this loss as mixed CE-2nd pass.
It turns out that mixed CE-2nd is better than CE and this is because $\tilde{y}$ resembles $\hat{y}$ (consider that $y^{\text{mix}}$ and $y$ have common parts).
However, it is still worse than mixed CE, indicating the importance of approximating $\hat{y}$.

\subsection{The Effect of $m$ in $\alpha_i$}
% {\color{blue} In this section, we explore how different choice of $\alpha_i$ influence the model performance. Other ablation studies can be found in the appendix.}

\begin{figure}
    \centering
    \includegraphics[scale=0.45]{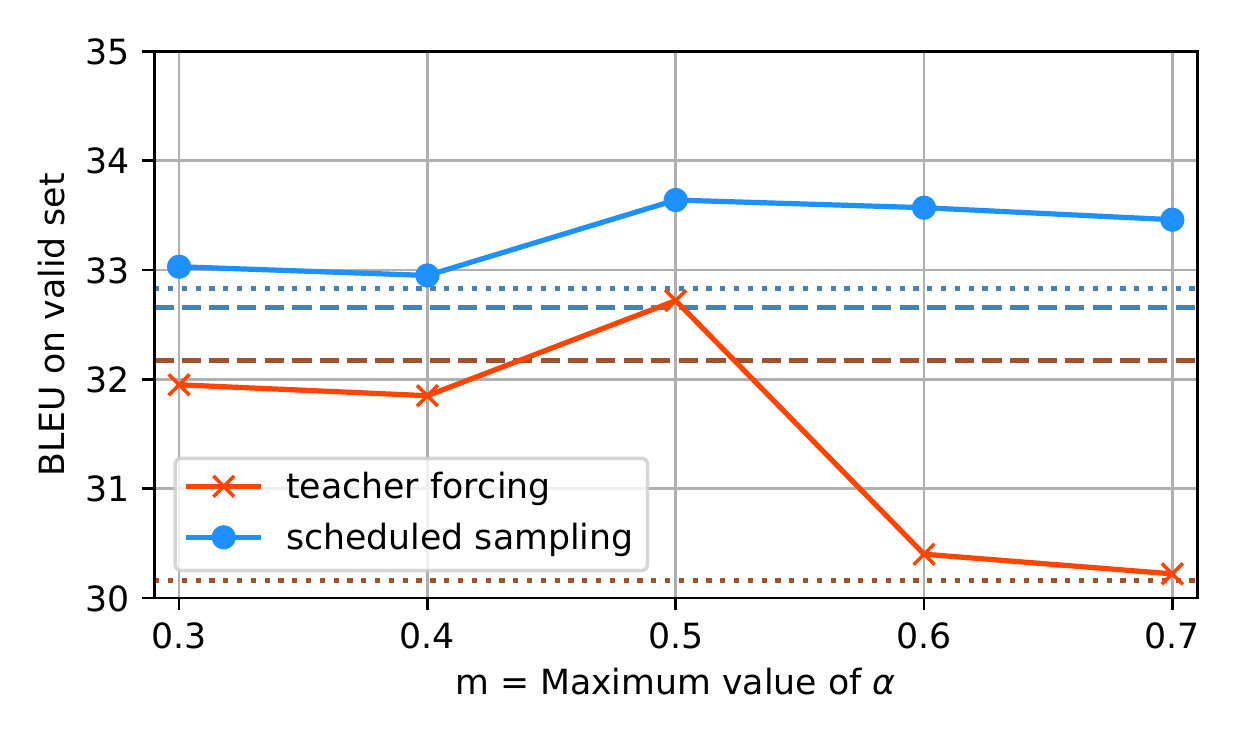}
    \caption{BLEU scores on the WMT'16 Ro-En validation set with different $m$ values. The blue and orange dotted lines denote the BLEU scores of the model with $\alpha_i=0.5$ while the dashed lines denote the result of training with CE loss.}
    \label{alpha}
    \vspace{-10pt}
\end{figure}

Here we tried different values of the constant $m= \{0.3,$ $0.4 , 0.5, 0.6, 0.7\}$ in Eq. \ref{eq3} in both teacher forcing and scheduled sampling training on WMT'16 Ro-En.
The best BLEU scores on validation set with different $m$ are shown in Fig. \ref{alpha}.
In teacher forcing, larger $m$ values favor the model's inconsistent predictions over the gold data whereas smaller values do not trust model predictions that much.
In scheduled sampling, larger $m$ values highlight the importance of reconciling two outputs in two passes while smaller $m$ values emphasize approximating the gold target.
In general, $m=0.5$ makes a good trade-off and the gap between different $m$ is smaller in scheduled sampling than in teacher forcing.
We also tried to fix $\alpha_i$ to be 0.5 throughout the training process and the best validation BLEU is shown in dotted lines in Fig. \ref{alpha}.
It turns out that this fixed strategy is more harmful in teacher forcing since it discourages the model from learning at the early stage when the model is still uninformative.
We also plotted the validation score of the models with CE loss in Fig. \ref{alpha} (dashed lines).
It can be seen that the fixed strategy still outperforms CE loss in scheduled sampling even though in teacher forcing CE exceeds the fixed strategy by a large margin.
%{More ablation studies can be found in the Appendix.}

\section{Related Work}

Neural machine translation has made significant progress in recent year \citep{cho-etal-2014-learning, NIPS2014_5346, DBLP:journals/corr/BahdanauCB14, NIPS2017_7181} 
But many auto-regressive models face exposure bias \citep{DBLP:journals/corr/RanzatoCAZ15}.
%Exposure bias has received much attention from the machine learning and NLP community. 
The mainstream solution to exposure bias is to train the model on its own predictions,  \citep{Daum2009SearchbasedSP,pmlr-v15-ross11a,Venkatraman2015ImprovingMP}.
% \citet{Daum2009SearchbasedSP} proposed a meta-algorithm SEARN, which constructs cost-sensitive examples from the model's own predictions and then trains a new model based on these examples.
% \citet{pmlr-v15-ross11a} later proposed a similar method called DAgger (Dataset Aggregation).
% \citet{Venkatraman2015ImprovingMP} presented DAD (Data As Demonstrator), a meta-algorithm, which also reuses the model's predictions for multi-step simulations. 
Later, \citet{NIPS2015_5956} proposed scheduled sampling for RNN training. 
There are also many variants of scheduled sampling \citep{Goyal2017DifferentiableSS, mihaylova-martins-2019-scheduled, zhang-etal-2019-bridging,duckworth2020parallel}.
% However, \citet{DBLP:journals/corr/Huszar15} argued that scheduled sampling can be problematic under certain circumstances. 
% Nevertheless, applying SS with MLE to auto-regressive models in conditional generation training often yields better performance \citep{Goyal2017DifferentiableSS, zhang-etal-2019-bridging}.  
In addition, \citet{DBLP:journals/corr/RanzatoCAZ15, DBLP:conf/iclr/BahdanauBXGLPCB17} incorporated the ideas in reinforcement learning into sequence prediction problem
% proposed that MIXER (Mixed Incremental Cross-Entropy Reinforce) algorithm which combines REINFORCE and cross-entropy loss to handle the large action spaces in language generation. 
% \citet{DBLP:conf/iclr/BahdanauBXGLPCB17} also borrowed ideas from reinforcement learning for sequence prediction. 
while \citet{wiseman-rush-2016-sequence,zhang-etal-2019-bridging} integrated beam search into the sequence-to-sequence training procedure. 
\citet{leblond2018searnn} linked RNNs with SEARN \citep{Daum2009SearchbasedSP} and proposed SEARNN. 
There are also some other interesting works addressing exposure bias, e.g., minimum risk training \citep{shen-etal-2016-minimum}, adversarial training \citep{NIPS2016_6099}. 
% \citet{goldberg-nivre-2012-dynamic} designed a dynamic oracle to provide a set of optimal actions in arc-eager dependency parsing.  
% \citet{shen-etal-2016-minimum} used  to mitigate exposure bias as well as loss-evaluation mismatch in machine translation. 
% \citet{NIPS2016_6099} presented an adversarial domain adaptation algorithm, professor forcing, to encourage the dynamics during training and test time to be the same.
More recently, \citet{schmidt-2019-generalization}, \citet{DBLP:journals/corr/abs-1905-10617} provided a new perspective to understand exposure bias.

Real noisy data problem in neural machine translation has been well studied \citep{koehn-etal-2018-findings, wang-etal-2018-denoising,belinkov2018synthetic,dakwale-monz-2019-improving}, but we've argued that NMT with clean data can also be treated as a noisy label problem in a sense (see Section \ref{noisy}).
\citet{song2020learning} provides a thorough review of approaches to handling noisy labels and one of the approaches called Bootstrapping \citep{43273} is similar to mixed CE.
However, there are several differences: 1) mixed CE is used in machine translation with clean labels; 2) when we treat those clean texts as ``noisy'' ones, the ratio of ``noisy'' labels is nearly 100\% while in \citet{43273} this number is much smaller; 3) mixed CE assigns linearly decay coefficients to the two {\color{black}log likelihood} with human priors while Bootstrapping selects a fixed value via cross-validation. 
\textcolor{black}{Another similar approach to mixed CE in teacher forcing is called self-knowledge distillation \citep{hahn-choi-2019-self} which uses the scaled Euclidean distance between the word embeddings of the target token and model's greedy prediction to compute the $\alpha$ values.
We re-implemented their approach but failed to find one proper scaled factor that could outperform the baseline in our setting. More details can be found in the Appendix.}
% \textcolor{red}{Another similar approach to mixed CE in teacher forcing is called self-knowledge distillation \citep{hahn-choi-2019-self} which uses the word embeddings to compute the $\alpha$ values but require a pre-training stage when the emebddings are still not informative.
% However, mixed CE in teacher forcing is motivated by the noisy label problem and we directly assign a linear decay value to $\alpha$ and thus avoid their pre-training step.
% Besides, we provide more in-depth analysis of mixed CE by comparing it to label smoothing and testing it on multiple references.}

\section{Conclusion}
In this paper, we propose mixed CE which can be used in teacher forcing training and scheduled sampling training in neural machine translation.
In teacher forcing training, mixed CE can make full use of the model's own predictions during training and tends to assign probability mass to the tokens related to the gold targets.
We systematically analyze the output distribution's properties of mixed CE and make a comparison with label smoothing.
In scheduled sampling, mixed CE forces the model to approximate not only the gold targets but also the greedy predictions in the first forward pass conditioned on gold inputs and thus mitigate exposure bias more effectively.
We demonstrate the effectiveness of mixed CE on several standard machine translation data sets at different scales, {\color{black}namely} WMT'16 Ro-En, WMT'16 Ru-En, WMT'14 En-De as well as two sets of additional challenging references.
Specifically, in a multi-reference set, mixed CE consistently outperforms CE across 10 additional references.
{\color{black}Such results demonstrate the effectiveness of our proposed mixed CE objective for neural machine translation.}
{\color{black} In the future, it would be also interesting to explore the use of mixed CE in non-autoregreesive machine translation and domain robustness problems.}

\section*{Acknowledgements}
We would like to thank the anonymous reviewers and the meta-reviewer for their insightful comments.
We would also like to thank Wen Zhang, Yang Feng, Zhanming Jie, Perry Lam for their help.
This research is  supported by Ministry of Education, Singapore, under its Academic Research Fund (AcRF) Tier 2 Programme (MOE AcRF Tier 2 Award No: MOE2017-T2-1-156). 
Any opinions, findings and conclusions or recommendations expressed in this material are those of the authors and do not reflect the views of the Ministry of Education, Singapore.


\bibliography{example_paper}
\bibliographystyle{icml2021}


%%%%%%%%%%%%%%%%%%%%%%%%%%%%%%%%%%%%%%%%%%%%%%%%%%%%%%%%%%%%%%%%%%%%%%%%%%%%%%%
%%%%%%%%%%%%%%%%%%%%%%%%%%%%%%%%%%%%%%%%%%%%%%%%%%%%%%%%%%%%%%%%%%%%%%%%%%%%%%%

\end{document}